\documentclass[11pt]{article}

\usepackage[preprint]{acl}

\usepackage{times}
\usepackage{latexsym}

\usepackage[T1]{fontenc}

\usepackage[utf8]{inputenc}

\usepackage{microtype}

\usepackage{inconsolata}

\usepackage{graphicx}
\usepackage{amsmath}
\usepackage{booktabs}   
\usepackage{multirow}   
\usepackage[table]{xcolor} 
\usepackage{makecell}
\usepackage{subcaption}
\usepackage{tabularx}
\usepackage{array}
\usepackage[most]{tcolorbox}
\usepackage{listings}
\definecolor{rubricred}{RGB}{180,55,45}
\definecolor{harnessgreen}{RGB}{45,120,70}
\definecolor{progressyellow}{RGB}{255,248,220}
\usepackage{soul}
\sethlcolor{progressyellow}

\definecolor{caseblue}{RGB}{235,242,250}
\definecolor{caseyellow}{RGB}{252,248,228}
\title{When the Critic Falls Behind: Co-Evolving Rubrics and Harnesses for Continual Agent Self-Improvement}

\author{
Hao Chen$^{1}$,
Yukun Yan$^{2}$\thanks{\hspace{1mm}Corresponding Authors.},
Ziyu Han$^{1}$,
Ye He$^{1}$,
Chuanzhe Guo$^{1}$, 
Shangzhan Li$^{1}$, \\
\textbf{Qi Shi$^{2}$,
Qingfu Zhu$^{1}$,
Maosong Sun$^{2}$,
Wanxiang Che$^{1}$\footnotemark[1]} \\
\textsuperscript{1}Research Center for Social Computing and Interactive Robotics, Harbin Institute of Technology \\
\textsuperscript{2}Department of Computer Science and Technology, Institute for AI, Tsinghua University \\
}

\begin{document}
\maketitle

\begin{abstract}
Autonomous agents increasingly depend on harnesses that govern interaction, memory, tools, and execution around large language models (LLMs). While recent work has begun to evolve these harnesses from execution experience, the supervision used to guide evolution often remains relatively static, making it difficult to track changing behavioral weaknesses across rounds and potentially limiting generalization to unseen tasks. To address this issue, we propose RuHarness, a rubric--harness co-evolution framework for continual agent self-improvement. RuHarness continually discovers and reassesses behavioral criteria from execution trajectories, prioritizes the most relevant unresolved weaknesses, and uses them to guide targeted harness updates while keeping the task objective fixed. Extensive experiments across diverse agent tasks and backbone models show that RuHarness outperforms strong harness-evolution baselines, particularly on unseen held-out tasks. Further analyses show that adaptive rubric prioritization tracks shifting weaknesses and supports more sustained improvement and stable generalization across evolution rounds.
\end{abstract}
\section{Introduction}
\begin{figure}[t]
\centering
\includegraphics[width=\linewidth]{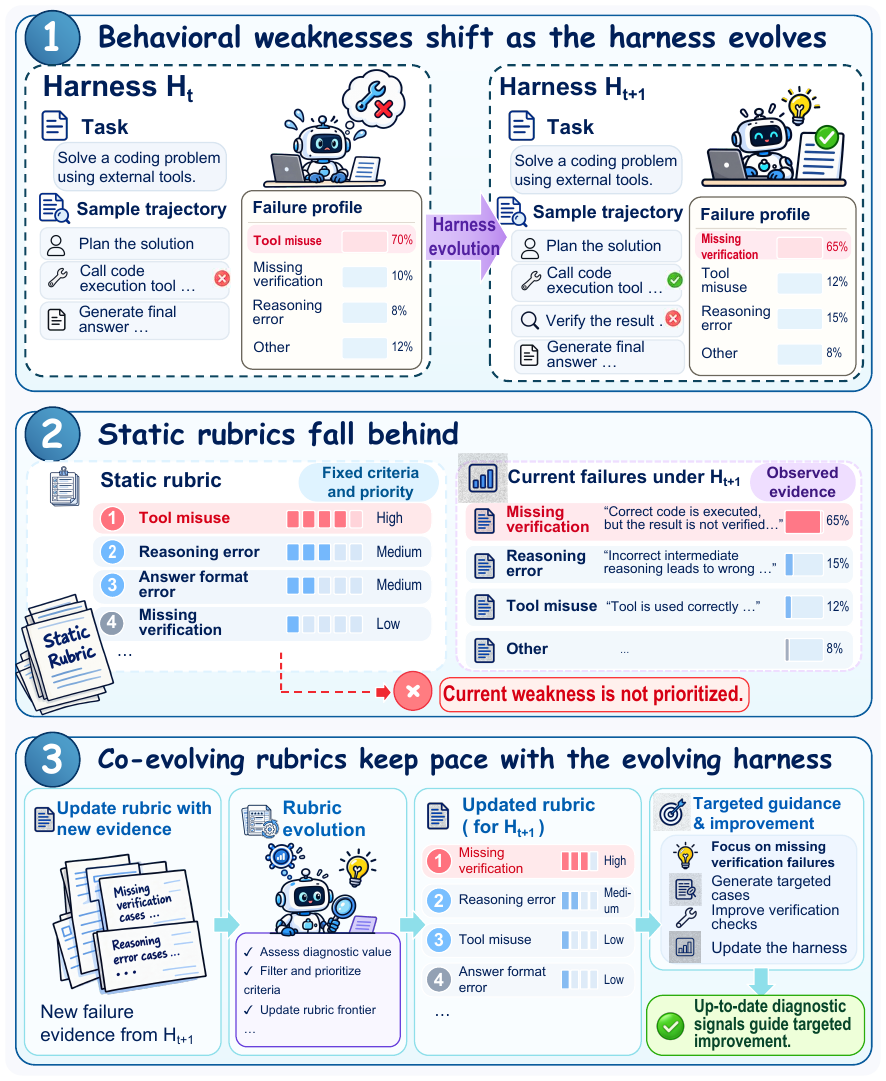}
\caption{Illustration of rubric–harness co-evolution. Static rubrics may fall behind as behavioral weaknesses shift during harness evolution, while co-evolving rubrics track emerging failures and guide harness improvement.}
\label{fig:introduction}
\end{figure}
As autonomous agents become increasingly prevalent, their performance depends not only on the capabilities of the underlying large language model (LLM) but also on the surrounding harness that governs interaction, memory, tools, and execution~\citep{luo2026harnessbank, pan2026evolvingagents}. Recent work has begun to evolve the harness itself from execution experience, enabling agents to improve without updating model parameters~\citep{zhang2026selfharness, lin2026harnessupdating, lee2026recursiveharnessselfimprovement}. While such approaches can yield substantial gains, recent controlled evaluations suggest that these improvements may weaken on unseen tasks~\citep{wang2026rethinking}, raising questions about whether harness optimization produces generalizable improvement.

One challenge behind this limited generalization is that repeated harness optimization may gradually overfit to task-specific failure patterns exposed during evolution~\citep{srikanth2026RSI}. Existing methods optimize harness against task-level feedback, which indicates whether the agent succeeds but provides limited information about the behavioral deficiencies underlying failure~\citep{zhang2026selfharness}. This limitation becomes increasingly important during continual evolution: as the harness improves, previously dominant failures may be mitigated, while subtler or previously unseen failure patterns emerge~\citep{nie2026tthe}. Figure~\ref{fig:introduction} illustrates that when the guidance used for evolution remains coarse-grained and relatively static, its discriminative and guiding value may gradually diminish~\citep{kim2025rethinkingreward,tian2026arco}. Consequently, the system lacks an explicit mechanism for tracking how the importance of different weaknesses changes throughout evolution. Continual harness evolution requires adaptive diagnosis that continually reassesses and reprioritizes behavioral weaknesses while keeping the underlying task objective fixed~\citep{dou2026agentslarge,zhang2026recursive}.

To this end, we propose RuHarness (\textbf{Ru}bric--\textbf{Harness} Co-Evolution), a framework for continual agent self-improvement through joint rubric and harness evolution. RuHarness continually derives behavioral criteria from execution trajectories, reassesses their relevance under the current harness, and prioritizes those that best capture unresolved weaknesses. The resulting rubric frontier guides targeted and coordinated updates to the corresponding harness components, while candidate harnesses are selected using task performance on an independent validation set. By adapting the diagnostic guidance as agent behavior changes while keeping the task objective fixed, RuHarness continually focuses harness optimization on the weaknesses that matter most at each stage of evolution.

Extensive experiments across diverse agent tasks and backbone models demonstrate that RuHarness consistently outperforms strong harness-evolution baselines, with particularly clear gains on unseen held-out tasks. The analyses also show that the rubric frontier adapts to shifting behavioral weaknesses, concentrates on diagnostically relevant criteria, and supports more sustained improvement across evolution rounds. These results indicate that co-evolving rubrics with the harness provides adaptive supervision for continual agent improvement, supporting both continued evolution across evolution rounds and stronger generalization.
\section{Related Work}
\textbf{Self-Evolving Agents.}
Agent self-evolution has gradually shifted from storing interaction experience to optimizing persistent external state. Reflection- and memory-based methods refine experience to improve future decisions~\citep{shinn2023reflexion,zhong2023memorybank}, while skill evolution treats reusable skills as editable knowledge that can be refined from execution feedback~\citep{zhu2026skillcoach,yang2026skillopt}. These paradigms extend naturally to the broader agent harness, where the optimization target expands from a single external component to the execution infrastructure, including prompts, tools, memory, and control logic~\citep{zhang2026scalinglaws}. Recent harness-evolution methods seek to diagnose failures from trajectories and iteratively revise this infrastructure~\citep{jiang2026harnessevolve}, with subsequent work broadening the editable space and making the evolution process more observable and verifiable~\citep{lee2026metaharness,lin2026AHE}. More broadly, recent studies have explored recursive self-improvement and harness--model co-evolution across repeated adaptation rounds~\citep{xia2026rrsi, yu2026coevolving, chen2026coharness,fan2026helix}. Yet these approaches still largely treat the diagnostic signals guiding evolution as fixed, leaving how such guidance should adapt to shifting behavioral weaknesses underexplored.

\textbf{Rubrics for Self-Improvement.}
Rubrics have evolved from structured evaluation criteria into a flexible interface between evaluation and optimization~\citep{chen2026holistic}. By decomposing complex objectives into explicit behavioral criteria, they provide fine-grained supervision and have increasingly been incorporated into reward modeling and reinforcement learning~\citep{gunjal2025rubrics,li2026rubricem}. More recent work makes the rubric itself adaptive: EvoLM~\citep{li2026evolm}, EvoRubrics~\citep{ding2026evorubrics}, and DynamicRubric~\citep{wang2026coevolving} co-evolve evaluative criteria with the optimized policy~\citep{guan2026evorubric}, while~\citet{wang2026serpo} extends this idea to a self-evolving loop over responses, rubrics, and policy. These studies suggest that effective supervision should evolve with the behavior it guides~\citep{jiang2026rubricarrow}. However, existing work mainly co-evolves rubrics with policies or responses~\citep{li2026stale}, leaving rubric adaptation for continual harness evolution largely unexplored.
\section{Method}
\begin{figure*}[t]
\centering
\includegraphics[width=\linewidth]{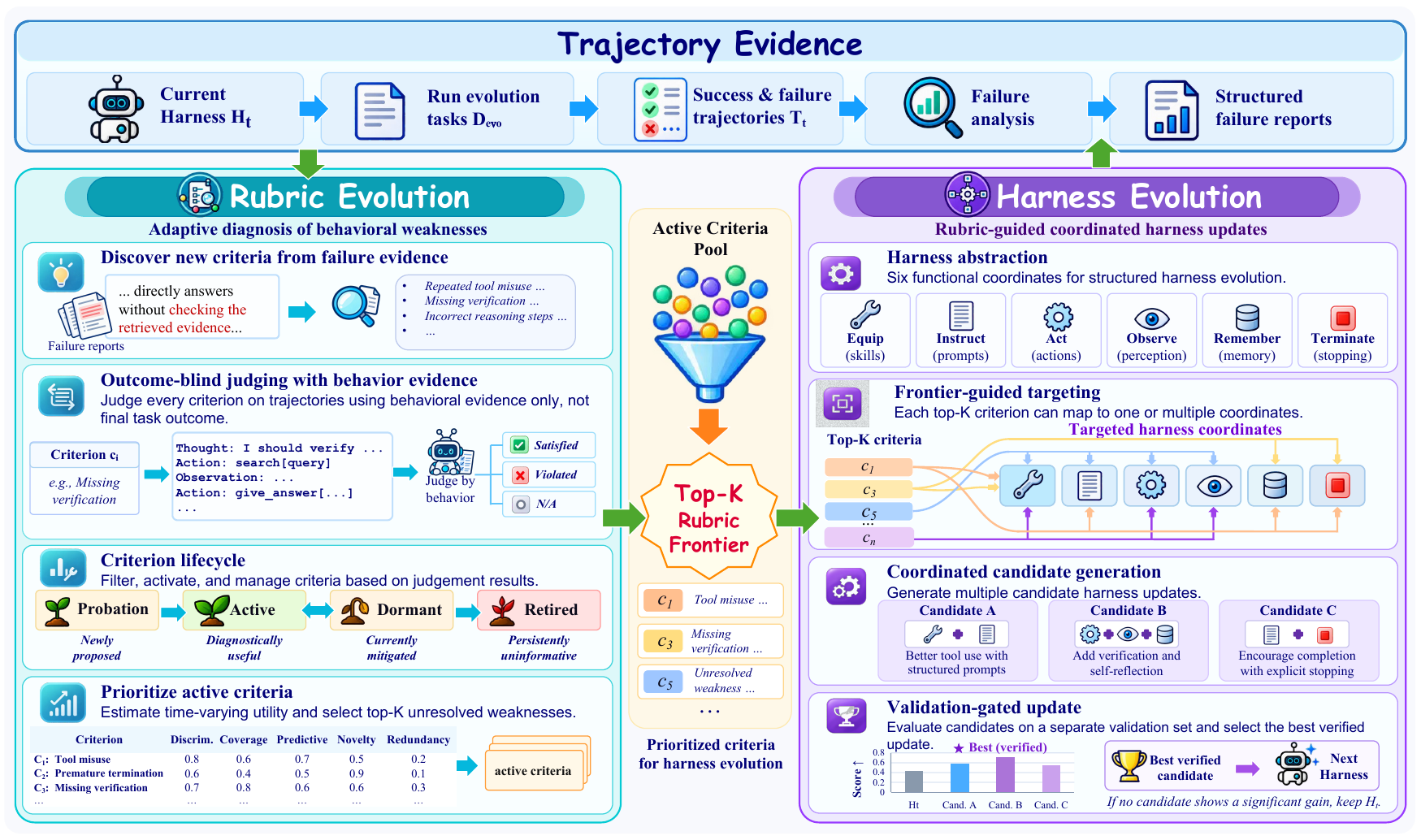}
\caption{Overview of the Rubric–Harness Co-Evolution Framework. Rubric evolution adaptively identifies and prioritizes unresolved behavioral weaknesses from trajectory evidence, while harness evolution performs targeted updates accordingly, yielding new behaviors that drive the next round of evolution.}
\label{fig:method}
\end{figure*}
In this section, we first present the overall design of RuHarness~(§\ref{3.1overall}), then describe how rubric evolution tracks the changing behavioral distribution~(§\ref{3.2RubricEvolution}) and guides subsequent harness evolution~(§\ref{3.3HarnessEvolution}), as illustrated in Figure~\ref{fig:method}.

\subsection{Overview of the Framework}\label{3.1overall}
RuHarness establishes a co-evolution process between the diagnostic rubric and the agent harness through iterative interaction. At evolution round \(t\), the system state is represented by \(\mathcal{S}_t=(H_t,R_t)\), where \(H_t\) denotes the current harness and \(R_t\) the maintained rubric pool. RuHarness updates the diagnostic signal from the behavior exposed by the current harness and uses the resulting signal to guide harness updates, thereby alternating between rubric evolution and harness evolution.

\textbf{Alternating evolution process.}
At each round, the current harness \(H_t\) is executed on a set of evolution tasks \(D_{\mathrm{evo}}\) to collect agent trajectories \(\mathcal{T}_t\). These trajectories are analyzed to identify remaining failure patterns, which drive the rubric update:
\begin{equation}
    R_{t+1} = \mathcal{E}_{R}\big(R_t,\mathcal{T}_t\big),
\end{equation}
where \(\mathcal E_R\) discovers new behavioral criteria and reassesses existing ones from \(\mathcal T_t\) detailed in §\ref{3.2RubricEvolution}. The updated rubric pool is then prioritized to obtain a set of high-priority criteria \(\widehat R_{t+1}\), which captures the behavioral weaknesses most relevant to the current harness. These criteria guide the rubric-conditioned harness update:
\begin{equation}
    H_{t+1} = \mathcal{E}_{H}\big(H_t,\widehat R_{t+1}\big),
\end{equation}
where \(\mathcal E_H\) denotes the harness evolution procedure detailed in §\ref{3.3HarnessEvolution}. The updated harness induces a new behavioral distribution in the next round, providing fresh evidence for subsequent rubric evolution and closing the co-evolution loop. The two processes are therefore coupled through agent behavior: the harness shapes the weaknesses that emerge, while the rubric determines which weaknesses are prioritized to guide subsequent evolution.

\subsection{Rubric Evolution: Adapting to Shifting Agent Weaknesses}\label{3.2RubricEvolution}

Given trajectories from \(H_t\), RuHarness summarizes each failed run into a structured failure report and aggregates these reports to identify recurring cross-task behavioral patterns. These patterns are converted into new criteria \(\mathcal{R}_{t+1}^{\mathrm{new}}\), which capture recurring weaknesses across tasks while remaining verifiable from trajectory evidence.

\textbf{Outcome-blind behavioral assessment.}
Newly discovered criteria are evaluated together with existing criteria on trajectories generated by the current harness. For each criterion \(c\) and trajectory \(\tau\), an outcome-blind judge assigns:
\begin{equation}
    J(c,\tau) \in \{\text{Satisfied}, \text{Violated}, \text{N/A}\},
\end{equation}
where the judge has access only to behavioral evidence in the trajectory and is blinded to the final task outcome. This prevents criterion assessment from directly leaking success or failure labels. Here, \textit{N/A} indicates that the behavior specified by the criterion does not occur, whereas \textit{Satisfied} indicates that it occurs and is handled correctly.

\textbf{Criterion admission and lifecycle.}
Based on these assessments, newly discovered criteria first undergo a \textit{birth check} stage before entering the maintained rubric pool. Criteria with insufficient diagnostic value are filtered at this stage, including those that are rarely applicable, fail to identify violations on unsuccessful trajectories, or trigger indiscriminately across outcomes. Criteria that pass this admission stage enter a four-state lifecycle: \textit{probation}, \textit{active}, \textit{dormant}, and \textit{retired}. Newly admitted criteria enter probation and are promoted to active once they exhibit sufficient diagnostic value. Active criteria become dormant when their associated failures are no longer observed, and may be reactivated upon recurrence, while persistently uninformative criteria are eventually retired. Only \textit{active} criteria are subsequently prioritized and used to guide harness evolution.

\textbf{Time-varying diagnostic utility.}
To determine which active criteria are most useful for the current harness, we assign each criterion a time-dependent diagnostic utility:
\begin{equation}
    \begin{aligned}
        U_{t+1}(c) ={}& \alpha D_{t+1}(c) + \beta C_{t+1}(c) + \gamma P_{t+1}(c) \\
        & + \delta N_{t+1}(c) - \lambda \mathrm{Red}_{t+1}(c),
    \end{aligned}
\end{equation}
where \(D_{t+1}(c)\) denotes \textit{discrimination} between successful and failed trajectories, \(C_{t+1}(c)\) measures the fraction of current failures \textit{covered} by the criterion, \(P_{t+1}(c)\) measures \textit{predictive validity} with respect to task failure, \(N_{t+1}(c)\) captures \textit{novel} failures identified by the criterion, and \(\mathrm{Red}_{t+1}(c)\) penalizes behavioral \textit{redundancy} with existing criteria. These statistics are derived from the preceding behavioral assessments and the corresponding task outcomes, and are recomputed for each harness state. Consequently, \(U_{t+1}(c)\) evolves with the harness: criteria associated with mitigated failures decrease in priority, while those capturing newly prominent failures receive higher utility.

\textbf{Adaptive rubric frontier.}
Since statistics can be uncertain under limited observations, we rank active criteria using a conservative posterior estimate of diagnostic utility. Specifically, we estimate a Dirichlet posterior from the joint task-outcome and criterion-assessment counts and use the lower $q$-quantile of the induced utility distribution:
\begin{equation}
    \widetilde{U}_{t+1}(c)
    =
    Q_q\!\left[p\!\left(U_{t+1}(c)\mid\mathcal{T}_t\right)\right],
\end{equation}
where $\widetilde{U}_{t+1}(c)$ denotes the conservative utility of criterion $c$, and $Q_q(\cdot)$ is the $q$-th posterior quantile. This reduces the influence of criteria whose high utility is supported by only a few observations. We select the top-$K$ criteria from the active rubric pool:
\begin{equation}
    \widehat{R}_{t+1}
    =
    \operatorname{TopK}_{c\in R_{t+1}^{\mathrm{active}}}
    \widetilde{U}_{t+1}(c),
\end{equation}
where $R_{t+1}^{\mathrm{active}}$ denotes the active criteria and $\widehat{R}_{t+1}$ the resulting rubric frontier.
The resulting frontier prioritizes unresolved behavioral weaknesses that remain both diagnostically informative and statistically reliable under the current harness.

Further details of rubric evolution are provided in Appendix~\ref{app:diagnostic_utility}--\ref{app:posterior_ranking}.

\subsection{Harness Evolution: Optimizing against the Moving Frontier}\label{3.3HarnessEvolution}
To structure the harness search space, we represent the editable harness using six functional coordinates: \texttt{Equip}, \texttt{Instruct}, \texttt{Act}, \texttt{Observe}, \texttt{Remember}, and \texttt{Terminate}. These coordinates respectively capture resource provisioning, behavioral instruction, action policy, observation processing, context management, and completion control. Each coordinate defines a corresponding set of editable components in the agent system.

\textbf{Frontier-guided targeting.}
Given the current rubric frontier \(\widehat{R}_{t+1}\), RuHarness identifies the harness coordinates that can directly address each prioritized behavioral weakness. The criterion-to-coordinate mapping is not constrained to be one-to-one, since a single failure pattern may involve multiple parts of the harness. Instead, each criterion may target multiple coordinates, allowing the corresponding interventions to be planned jointly.

\textbf{Coordinated candidate generation.}
Based on these criterion-to-coordinate assignments, RuHarness jointly plans a small set of candidate harness updates. Each candidate specifies a concrete hypothesis for addressing the diagnosed weaknesses and may modify multiple related coordinates jointly. We represent each candidate as a coordinated set of coordinate-level modifications:
\begin{equation}
    \Delta_j
    =
    \left\{
    \delta_j^{(k)}
    \right\}_{k\in M_j},
\end{equation}
where $M_j$ is the set of modified coordinates and $\delta_j^{(k)}$ denotes the corresponding modification. The coordinated modifications are applied to the current harness through the update operator $\oplus$, yielding the candidate harness:
\begin{equation}
    H_t^{(j)} = H_t \oplus \Delta_j.
\end{equation}

Each candidate is treated as a single multi-coordinate update, preserving the interactions among jointly planned modifications. Before evaluation, each candidate undergoes a lightweight verification step to check whether the proposed modifications are consistent with the diagnosed weaknesses and supporting evidence. The verifier only accepts or rejects the candidate and does not rewrite it or estimate task performance.

\textbf{Validation-gated update.}
Candidates that pass verification are evaluated together with the current harness on a separate set of validation tasks \(D_{\mathrm{val}}\), used exclusively for candidate selection and excluded from both evolution and final evaluation. This separation decouples candidate selection from the trajectories used to drive evolution. Let \(S(H,D_{\mathrm{val}})\) denote the average validation score of harness \(H\). Let \(\mathcal{H}_t^{\mathrm{cand}}\) denote the set of candidate harnesses generated at round \(t\). The candidate with the highest validation score is selected as:
\begin{equation}
    H_t^\star
    \in
    \arg\max_{H\in\mathcal{H}_t^{\mathrm{cand}}}
    S(H,D_{\mathrm{val}}),
\end{equation}
and commit the update only if its improvement over the current harness exceeds a threshold \(\epsilon\):
\begin{equation}
    H_{t+1}
    =
    \begin{cases}
        H_t^\star,
        & S(H_t^\star,D_{\mathrm{val}})
        - S(H_t,D_{\mathrm{val}}) > \epsilon,\\
        H_t,
        & \text{otherwise}.
    \end{cases}
\end{equation}
A candidate is advanced to the next evolution round only if its validation improvement exceeds the acceptance threshold, with the current harness retained otherwise.
\section{Experimental Methodology}
This section presents our experimental setup, covering models and environment, benchmarks, metrics, baselines, and implementation details.

\textbf{Models and Environment.}
We conduct experiments with Qwen3.5-35B-A3B~\citep{qwen35blog} and GLM-4.7-Flash~\citep{5team2025glm45} as the backbone models. The agent is instantiated with DeepAgent~\citep{li2026deepagent}, which provides the basic execution framework for tool use, filesystem interaction, and multi-step task solving. Based on this framework, we construct the initial harness and expose its configurable components for subsequent evolution, while keeping the underlying model fixed throughout the experiments. All stages of task execution and evolution are performed using the corresponding backbone model itself.

\textbf{Benchmarks and Metrics.}
Evaluation is conducted on EvoAgentBench~\citep{gao2026evoagentbench}, a multi-domain benchmark for long-horizon agent self-evolution. It covers five agent domains: information retrieval based on BrowseComp-Plus~\citep{chen2025browsecompplus}, mathematical reasoning based on OmniMath~\citep{gao2024omnimath}, software engineering based on SWE-Bench~\citep{jimenez2024swebench}, code implementation based on LiveCodeBench~\citep{jain2024livecodebench}, and knowledge work based on GDPVal~\citep{patwardhan2025gdpval}. We use Pass@1 as the primary metric and evaluate each domain with its benchmark evaluator~\citep{tang2026memoryskill}.

\textbf{Baselines.}
For comprehensive comparison, we compare RuHarness against representative self-evolving agent methods that optimize different external states. ACE evolves persistent contextual knowledge from interaction experience~\citep{zhang2026ace}, while EvoSkill discovers and refines reusable skills through failure analysis~\citep{alzubi2026evoskill}. For harness evolution, we include Self-Harness, which mines model-specific weaknesses and proposes targeted updates~\citep{zhang2026selfharness}; Meta-Harness, which searches executable harness code using accumulated candidates and trajectories~\citep{lee2026metaharness}; and Agentic Harness Engineering (AHE), which improves harnesses through observable and verifiable component updates~\citep{lin2026AHE}. Together, these baselines cover context, skill, and full-harness evolution.

\textbf{Implementation Details.}
We use disjoint evolution (\emph{held-in}) and validation splits, with the test set (\emph{held-out}) reserved for final reporting only. At each round, RuHarness generates up to three coordinated harness candidates and accepts at most one update when its validation improvement exceeds $\epsilon=0.05$. For diagnostic utility, we set $\alpha=\gamma=1.0$ and $\beta=\delta=\lambda=0.5$. Rubric guidance is obtained by ranking criteria with the $q$-th posterior quantile ($q=0.2$) and selecting the Top-$K$ criteria with $K=10$. Each candidate modifies at most three harness components. More experimental details are provided in the Appendix~\ref{app:experimental_details}.

\section{Results and Analysis}
In this section, we first evaluate the performance of RuHarness and conduct ablation studies. We then analyze the evolution of the rubric frontier and how adaptive prioritization guides harness improvement. Finally, we compare evolution trajectories to assess sustained improvement and generalization.

\begin{table*}[t]
\centering
\small
\setlength{\tabcolsep}{1.3pt}
\renewcommand{\arraystretch}{1.10}

\begin{tabularx}{\textwidth}{
@{}
>{\raggedright\arraybackslash}p{0.17\textwidth}
*{12}{>{\centering\arraybackslash}X}
@{}
}
\toprule

\multirow{2}{*}{\textbf{Method}}
& \multicolumn{2}{c}{\makecell{\textbf{Information}\\\textbf{Retrieval}}}
& \multicolumn{2}{c}{\makecell{\textbf{Knowledge}\\\textbf{Work}}}
& \multicolumn{2}{c}{\makecell{\textbf{Software}\\\textbf{Engineering}}}
& \multicolumn{2}{c}{\makecell{\textbf{Code}\\\textbf{Implementation}}}
& \multicolumn{2}{c}{\makecell{\textbf{Math}\\\textbf{Reasoning}}}
& \multicolumn{2}{c}{\textbf{Average}} \\

\cmidrule(lr){2-3}
\cmidrule(lr){4-5}
\cmidrule(lr){6-7}
\cmidrule(lr){8-9}
\cmidrule(lr){10-11}
\cmidrule(lr){12-13}

& {\scriptsize Held-in} & {\scriptsize Held-out}
& {\scriptsize Held-in} & {\scriptsize Held-out}
& {\scriptsize Held-in} & {\scriptsize Held-out}
& {\scriptsize Held-in} & {\scriptsize Held-out}
& {\scriptsize Held-in} & {\scriptsize Held-out}
& {\scriptsize Held-in} & {\scriptsize Held-out} \\

\midrule

\rowcolor{gray!12}
\multicolumn{13}{c}{\textit{\textbf{Qwen3.5-35B-A3B}}} \\

Vanilla Agent
& 28.0 & 20.0
& 51.7 & 40.0
& 40.0 & 50.0
& 83.6 & 65.1
& 69.5 & 46.0
& 54.6 & 44.2 \\

ACE~{\scriptsize(\citeyear{zhang2026ace})}
& 31.0 & 21.5
& 56.7 & 40.0
& 46.7 & 55.6
& 86.7 & \textbf{69.8}
& 67.9 & 45.0
& 57.8 & 46.4 \\

EvoSkill~{\scriptsize(\citeyear{alzubi2026evoskill})}
& \underline{36.0} & \underline{27.7}
& 53.3 & 42.0
& 44.4 & 47.2
& 85.9 & \underline{68.6}
& 70.3 & 47.0
& 58.0 & 46.5 \\

Self-Harness~{\scriptsize(\citeyear{zhang2026selfharness})}
& 34.0 & \underline{27.7}
& 53.3 & 44.0
& \underline{51.1} & 55.6
& \textbf{90.6} & 67.4
& \textbf{72.9} & 47.0
& 60.4 & 48.3 \\

AHE~{\scriptsize(\citeyear{lin2026AHE})}
& 33.0 & 24.6
& 55.0 & \underline{50.0}
& \textbf{53.3} & \underline{58.3}
& 88.3 & 66.3
& 71.1 & 47.0
& 60.1 & \underline{49.2} \\

Meta-Harness~{\scriptsize(\citeyear{lee2026metaharness})}
& 34.0 & 26.2
& \underline{58.3} & 36.0
& \underline{51.1} & 52.8
& 89.1 & \underline{68.6}
& 71.1 & \underline{49.0}
& \underline{60.7} & 46.5 \\

\rowcolor{blue!15}
\textbf{RuHarness (ours)}
& \textbf{37.0} & \textbf{30.8}
& \textbf{60.0} & \textbf{52.0}
& \textbf{53.3} & \textbf{63.9}
& \underline{89.8} & \textbf{69.8}
& \underline{72.1} & \textbf{52.0}
& \textbf{62.4} & \textbf{53.7} \\

\midrule

\rowcolor{gray!12}
\multicolumn{13}{c}{\textit{\textbf{GLM-4.7-Flash}}} \\

Vanilla Agent
& 13.0 & 6.2
& 45.0 & 32.0
& 31.1 & 33.3
& 49.2 & 43.0
& 29.5 & 8.0
& 33.6 & 24.5 \\

ACE~{\scriptsize(\citeyear{zhang2026ace})}
& 14.0 & 4.6
& 53.3 & 42.0
& 37.8 & 44.4
& 54.7 & 47.7
& 26.8 & 11.0
& 37.3 & 29.9 \\

EvoSkill~{\scriptsize(\citeyear{alzubi2026evoskill})}
& 10.0 & \underline{9.2}
& 50.0 & 36.0
& 44.4 & 38.9
& 50.0 & 46.5
& 29.2 & 7.0
& 36.7 & 27.5 \\

Self-Harness~{\scriptsize(\citeyear{zhang2026selfharness})}
& 15.0 & \underline{9.2}
& 50.0 & 44.0
& \underline{46.7} & \underline{47.2}
& \textbf{61.7} & \underline{51.2}
& 29.7 & \underline{12.0}
& \underline{40.6} & \underline{32.7} \\

AHE~{\scriptsize(\citeyear{lin2026AHE})}
& 15.0 & 7.7
& 46.7 & 36.0
& 42.2 & 41.7
& 60.2 & 50.0
& 29.2 & 10.0
& 38.7 & 29.1 \\

Meta-Harness~{\scriptsize(\citeyear{lee2026metaharness})}
& \underline{16.0} & 7.7
& \underline{56.7} & \underline{46.0}
& 40.0 & 38.9
& 59.4 & 47.7
& \underline{30.0} & \textbf{15.0}
& 40.4 & 31.1 \\

\rowcolor{blue!15}
\textbf{RuHarness (ours)}
& \textbf{18.0} & \textbf{13.8}
& \textbf{58.3} & \textbf{56.0}
& \textbf{48.9} & \textbf{52.8}
& \underline{60.9} & \textbf{52.3}
& \textbf{30.5} & \textbf{15.0}
& \textbf{43.3} & \textbf{38.0} \\

\bottomrule
\end{tabularx}

\caption{
Performance comparison across five domains on the held-in and held-out sets.
Average denotes the macro-average across all five domains.
The best result is shown in \textbf{bold}, and the second-best result is \underline{underlined}.
}
\label{tab:main_results}
\end{table*}

\subsection{Overall Performance}
We evaluate RuHarness across five agent domains on EvoAgentBench using Qwen3.5-35B-A3B and GLM-4.7-Flash, as shown in Table~\ref{tab:main_results}. Overall, RuHarness outperforms all the baselines, achieving the best average performance on both held-in and held-out sets across the two backbone models.

\textbf{Overall Effectiveness.}
On Qwen3.5-35B-A3B, RuHarness improves the strongest baseline by 1.7 points on held-in and 4.5 points on held-out. The gains are even larger on GLM-4.7-Flash, reaching 2.7 and 5.3 points, respectively. These results indicate that rubric--harness co-evolution provides stable improvements beyond a specific backbone.

\textbf{Generalization to Unseen Tasks.}
The advantage of RuHarness is more pronounced on the held-out sets. Although several baselines achieve competitive or even better results on individual held-in tasks, RuHarness attains the highest average held-out performance for both models and exhibits a smaller held-in--held-out gap. This suggests that its improvements are less tied to the evolution tasks and transfer more effectively to unseen tasks.

\textbf{Robustness across Tasks and Models.}
RuHarness performs consistently well across heterogeneous tasks and both backbone models, achieving the strongest average held-out performance and competitive results across individual domains. This suggests that its gains are not tied to a specific task type or backbone.

\subsection{Ablation Study}
\begin{table*}[t]
\centering
\small
\setlength{\tabcolsep}{1.6pt}
\renewcommand{\arraystretch}{1.08}

\begin{tabular*}{\textwidth}{@{\extracolsep{\fill}}l cc cc cc cc cc cc@{}}
\toprule
\multirow{2}{*}{\textbf{Method}}
& \multicolumn{2}{c}{\makecell{\textbf{Information}\\\textbf{Retrieval}}}
& \multicolumn{2}{c}{\makecell{\textbf{Knowledge}\\\textbf{Work}}}
& \multicolumn{2}{c}{\makecell{\textbf{Software}\\\textbf{Engineering}}}
& \multicolumn{2}{c}{\makecell{\textbf{Code}\\\textbf{Implementation}}}
& \multicolumn{2}{c}{\makecell{\textbf{Math}\\\textbf{Reasoning}}}
& \multicolumn{2}{c}{\textbf{Average}} \\
\cmidrule(lr){2-3}
\cmidrule(lr){4-5}
\cmidrule(lr){6-7}
\cmidrule(lr){8-9}
\cmidrule(lr){10-11}
\cmidrule(lr){12-13}

& {\scriptsize Held-in} & {\scriptsize Held-out}
& {\scriptsize Held-in} & {\scriptsize Held-out}
& {\scriptsize Held-in} & {\scriptsize Held-out}
& {\scriptsize Held-in} & {\scriptsize Held-out}
& {\scriptsize Held-in} & {\scriptsize Held-out}
& {\scriptsize Held-in} & {\scriptsize Held-out} \\
\midrule

\textbf{RuHarness}
& \textbf{37.0} & \textbf{30.8}
& \textbf{60.0} & \textbf{52.0}
& \textbf{53.3} & \textbf{63.9}
& \textbf{89.8} & \textbf{69.8}
& 72.1 & \textbf{52.0}
& \textbf{62.4} & \textbf{53.7} \\

w/o Rubric Module
& 31.0 & 23.1
& 51.7 & 44.0
& 48.9 & 52.8
& 82.8 & 65.1
& 71.1 & 44.0
& 57.1 & 45.8 \\

w/o Rubric Evolution
& 35.0 & 29.2
& 55.0 & 34.0
& 44.4 & 55.6
& 76.6 & 57.0
& 71.3 & 46.0
& 56.5 & 44.4 \\

w/o Rubric Frontier
& 35.0 & 30.8
& 56.7 & 50.0
& 46.7 & 50.0
& 84.4 & 67.4
& \textbf{72.6} & 47.0
& 59.1 & 49.0 \\

\bottomrule
\end{tabular*}
\caption{
Ablation study of RuHarness on Qwen3.5-35B-A3B across five domains.
}
\label{tab:ablation}
\end{table*}
We further examine the contribution of key components of RuHarness, as summarized in Table~\ref{tab:ablation}.

\textbf{Necessity of the Rubric Module.}
Removing the rubric module lowers the average held-in score from 62.4 to 57.1 and the held-out score from 53.7 to 45.8. This indicates that directly evolving the harness from task feedback and trajectories provides insufficient behavioral diagnosis, highlighting the role of the rubric in translating task outcomes into actionable weaknesses.

\textbf{Effect of Rubric Evolution.}
To isolate the effect of rubric evolution, we retain the initial rubric as a fixed diagnostic rubric throughout the evolution process. Surprisingly, this variant performs even worse than removing the rubric module entirely, suggesting that a static rubric can become counterproductive as the harness evolves. When previously dominant failures are mitigated, outdated criteria may continue to steer optimization toward less relevant behaviors. Further analysis of this experimental process is provided in Appendix~\ref{app:dynamic_rubric_generalization}.

\textbf{Role of the Rubric Frontier.}
Using the entire ranked rubric pool instead of the selected frontier also leads to performance degradation. This suggests that more rubrics do not necessarily provide better guidance, as focusing on relevant criteria reduces interference from low-value signals and yields more effective harness updates. A more detailed examination is provided in Appendix~\ref{app:frontier_rubric}.

\subsection{Analysis of Adaptive Rubric Tracking}
\begin{figure}[t]
    \centering

    \begin{subfigure}[b]{0.49\columnwidth}
        \centering
        \includegraphics[width=\linewidth]{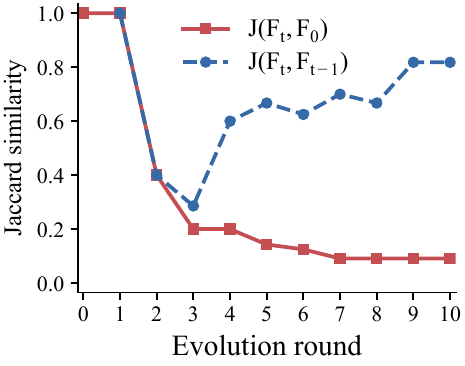}
        \caption{Qwen3.5 / Code}
        \label{fig:qwen_code_similarity}
    \end{subfigure}
    \hfill
    \begin{subfigure}[b]{0.49\columnwidth}
        \centering
        \includegraphics[width=\linewidth]{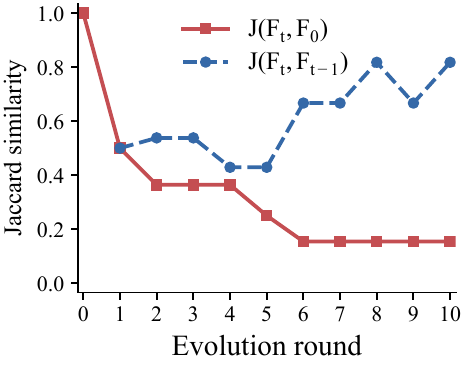}
        \caption{GLM-4.7 / SWE}
        \label{fig:glm_swe_similarity}
    \end{subfigure}

    \caption{Rubric similarity across evolution rounds.}
    \label{fig:rubric_similarity}
\end{figure}
To examine how the diagnostic focus changes throughout harness evolution, we measure the Jaccard similarity between rubric frontiers across rounds. Specifically, each frontier is represented by its selected criterion IDs, and we compare the frontier at round \(t\) with both the initial frontier \(F_0\) and the frontier from the preceding round \(F_{t-1}\). We conduct this analysis on two representative settings, Qwen3.5-35B-A3B on Code Implementation and GLM-4.7-Flash on Software Engineering.

As shown in Figure~\ref{fig:rubric_similarity}, the similarity to the initial frontier decreases sharply in the early rounds and remains low thereafter. By contrast, the similarity between consecutive frontiers first declines, then gradually increases and stabilizes at a relatively high level. These two trends reveal a characteristic evolution pattern: the diagnostic frontier progressively departs from its initial configuration, while changes between adjacent rounds become increasingly gradual. This behavior suggests that early harness updates substantially reshape the set of weaknesses that require attention, whereas later evolution proceeds through more localized adjustments. Overall, the results confirm that the behavioral weaknesses most relevant to harness improvement are not static, supporting the need for an evolving rubric that continuously tracks changes in the agent's failure landscape. A finer-grained visualization of criterion-level state transitions across evolution rounds is provided in Figure~\ref{fig:rubric_lifecycle}.

\subsection{Effectiveness of Adaptive Rubric Prioritization on Harness Evolution}
\begin{figure}[t]
    \centering

    \begin{subfigure}[t]{0.98\columnwidth}
        \centering
        \includegraphics[width=\linewidth]{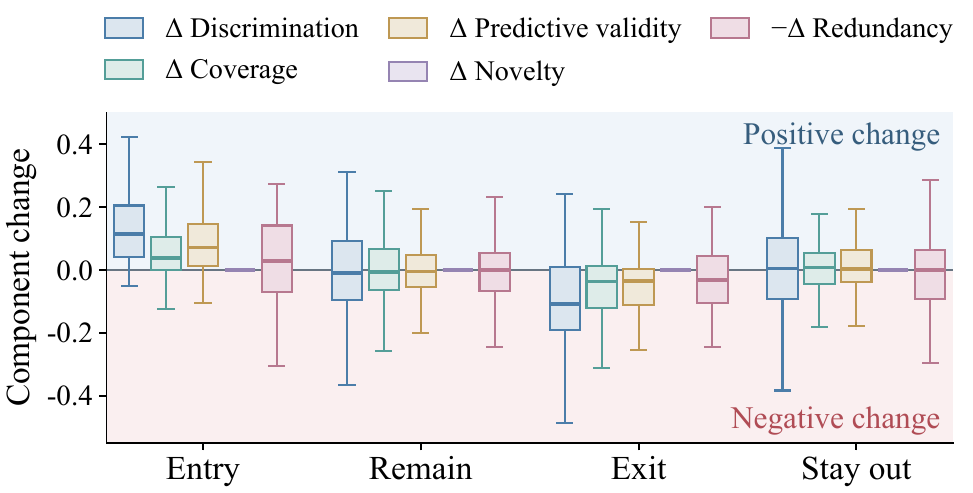}
        \caption{Diagnostic changes across frontier transitions.}
        \label{fig:diagnostic_transition}
    \end{subfigure}

    \vspace{0.4em}

    \begin{subfigure}[t]{0.48\columnwidth}
        \centering
        \includegraphics[width=\linewidth]{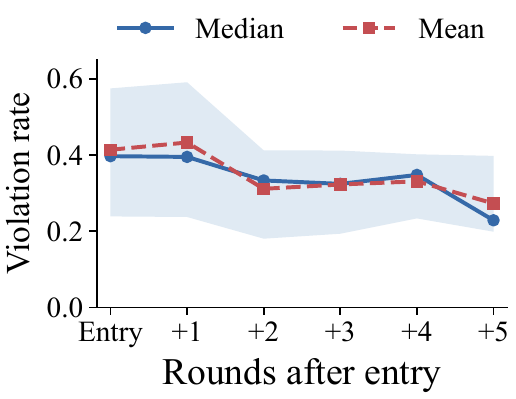}
        \caption{Violation rates after criteria become prioritized.}
        \label{fig:post_entry_violation}
    \end{subfigure}
    \hfill
    \begin{subfigure}[t]{0.48\columnwidth}
        \centering
        \includegraphics[width=\linewidth]{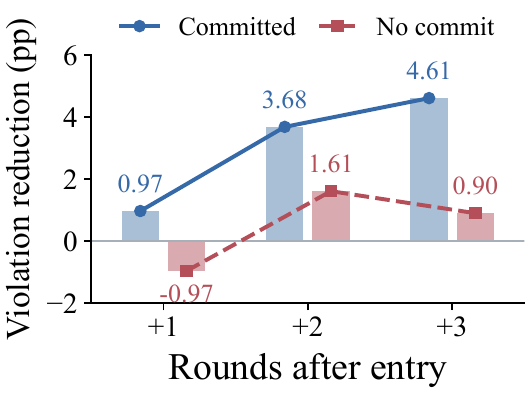}
        \caption{Violation reduction with and without harness updates.}
        \label{fig:fixed_cohort_reduction}
    \end{subfigure}

    \caption{
    Analysis of adaptive rubric prioritization and subsequent harness improvement.
    }
    \label{fig:frontier_analysis}
\end{figure}
To examine whether adaptive rubric prioritization translates into subsequent behavioral improvement, we analyze criterion-level dynamics around frontier transitions and after frontier entry. Specifically, we first compare how the diagnostic components of criteria change when they enter, remain in, exit, or stay outside the frontier. We then align newly prioritized criteria at their frontier-entry round and track how their violation rates evolve over subsequent rounds. To further relate this trend to harness evolution, we additionally distinguish cases with and without committed harness updates.

\textbf{Frontier transitions track diagnostic relevance.}
To characterize how diagnostic relevance changes with frontier membership, we group criteria by their transitions between two adjacent evolution rounds: entering the frontier, remaining in the frontier, exiting the frontier, or staying outside the frontier. For each group, we measure the round-to-round changes in the diagnostic utility components defined in Section~\ref{3.2RubricEvolution}, as shown in Figure~\ref{fig:diagnostic_transition}. As expected, criteria entering the frontier show increases in discrimination, coverage, and predictive validity together with reduced redundancy, while exiting criteria exhibit the opposite pattern. Criteria that remain inside or outside the frontier change substantially less. These results confirm that frontier turnover reflects changes in criterion-level diagnostic relevance rather than arbitrary replacement.

\textbf{Prioritized weaknesses are progressively mitigated.}
To assess whether frontier prioritization translates into subsequent behavioral improvement, Figure~\ref{fig:post_entry_violation} examines what happens after a criterion enters the frontier. The violation rate remains nearly unchanged in the first round after entry, but decreases noticeably over the following rounds. This delayed reduction suggests that newly prioritized weaknesses are not necessarily resolved immediately, but are progressively mitigated as subsequent harness evolution proceeds. The trend therefore links frontier prioritization with later behavioral improvement, suggesting that the evolving rubric does more than reorganize diagnostic signals and helps focus subsequent optimization on weaknesses that can be progressively reduced.

\textbf{Committed harness updates yield stronger longer-horizon mitigation.}
To more directly relate behavioral improvement to harness evolution, Figure~\ref{fig:fixed_cohort_reduction} compares frontier-entry criteria under committed and non-committed harness updates, where \emph{committed} denotes a candidate update that passes the validation gate and is adopted for subsequent evolution. The committed group shows a clear and sustained increase in violation reduction over subsequent rounds, whereas the non-committed group fluctuates without a consistent improvement trend. This contrast indicates that frontier-guided prioritization is more likely to translate into persistent behavioral gains when the proposed harness update is successfully validated and adopted.

Taken together, these results show that adaptive rubric prioritization tracks changing diagnostic relevance and helps focus subsequent harness evolution on weaknesses that are progressively mitigated.
\begin{figure}[t]
    \centering

    \begin{subfigure}[b]{0.49\columnwidth}
        \centering
        \includegraphics[width=\linewidth]{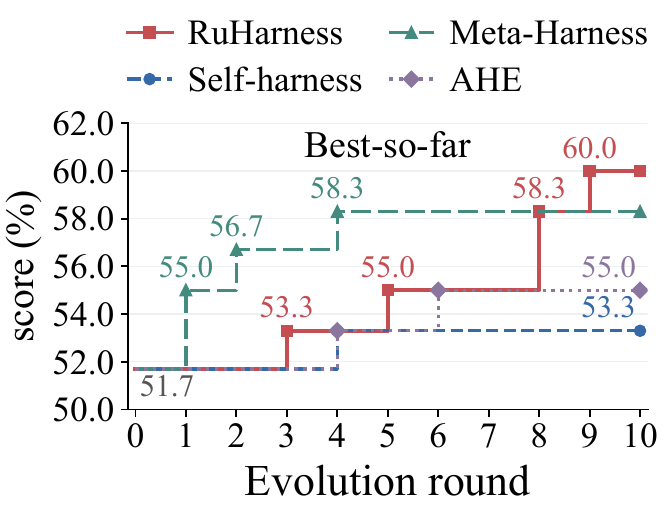}
        \caption{Held-in Task}
        \label{fig:evolution_hold_in}
    \end{subfigure}
    \hfill
    \begin{subfigure}[b]{0.49\columnwidth}
        \centering
        \includegraphics[width=\linewidth]{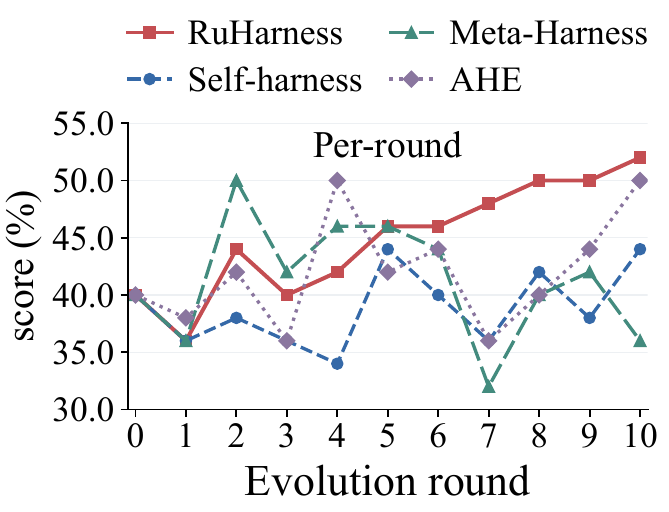}
        \caption{Held-out Task}
        \label{fig:evolution_hold_out}
    \end{subfigure}

    \caption{Evolution trajectories of different methods.}
    \label{fig:evolution_comparsion}
\end{figure}

\subsection{Evolution Progress and Generalization}
To examine performance across evolution rounds, Figure~\ref{fig:evolution_comparsion} compares the evolution trajectories of different harness evolution methods on the Qwen3.5-35B-A3B model with the knowledge work task. We report best-so-far held-in performance to reflect cumulative progress, while held-out performance is evaluated at each round to track generalization.

As shown in Figure~\ref{fig:evolution_hold_in}, on the held-in evolution tasks, several baselines improve quickly in early rounds but soon plateau, whereas RuHarness continues to gain in later rounds and achieves the highest performance. This suggests that dynamically updating the diagnostic focus helps expose new optimization opportunities after previously dominant weaknesses have been addressed.

The held-out trajectories in Figure~\ref{fig:evolution_hold_out} show that baseline performance fluctuates more substantially across rounds, while RuHarness exhibits a more stable upward trend in later stages. This suggests that the improvements acquired during evolution transfer more consistently beyond the held-in tasks. Together, these results indicate that adaptive rubric guidance supports both sustained improvement on held-in tasks and more stable generalization to held-out tasks throughout the evolution process.
\section{Conclusion}
This paper presents RuHarness, a rubric--harness co-evolution framework for continual agent self-improvement. RuHarness dynamically adapts diagnostic criteria to the agent's evolving behavioral weaknesses and uses the resulting rubric frontier to guide targeted harness optimization. Extensive experiments across diverse tasks and backbone models show that RuHarness achieves strong gains in both held-in performance and generalization to unseen held-out tasks. Further analyses show that the rubric frontier tracks shifting diagnostic relevance and supports sustained harness improvement over repeated evolution rounds. These results highlight the importance of jointly adapting diagnostic supervision and the agent harness for more sustained and generalizable agent evolution.

\section*{Limitations}
Although RuHarness is evaluated across multiple tasks and backbone models, all experiments are conducted within a single underlying agent framework. Evaluating the six-coordinate harness abstraction on a broader range of agent architectures would further establish its generality. In addition, due to computational resource constraints, our experiments are conducted on benchmark tasks under a fixed evolution horizon. Examining longer-term and more open-ended adaptation settings would provide a broader view of the resulting co-evolution dynamics.


\bibliography{main}
\clearpage
\newpage
\appendix

\section{Appendix}\label{sec:appendix}

\subsection{Experimental Details}
\label{app:experimental_details}

In this subsection, we provide the dataset splits, harness abstraction and editable interfaces, and prompt templates used throughout the experiments.

\textbf{Dataset Statistics and Settings.}
We conduct experiments on five agent tasks derived from EvoAgentBench~\citep{gao2026evoagentbench}, covering information retrieval, knowledge work, software engineering, code implementation, and mathematical reasoning. EvoAgentBench reorganizes existing benchmarks into train and test sets to evaluate whether improvements acquired during evolution transfer to unseen tasks. For our experiments, we further split the original training set into evolution and validation subsets, while preserving the original task distribution. The evolution subset is used to drive rubric--harness evolution, whereas the validation subset is used exclusively for candidate selection. We also remove a small number of tasks that cannot be reliably executed in our experimental environment. The resulting dataset statistics are summarized in Table~\ref{tab:dataset_statistics}.

We report performance on the evolution split as \emph{held-in} performance. The original test split is never accessed during evolution or candidate selection and is reserved exclusively for \emph{held-out} evaluation. All splits remain fixed throughout the experiments and are kept identical across all baseline methods to ensure a fair comparison. Evolution rounds, candidate budgets, and validation-based selection protocols are matched across harness-evolution methods unless otherwise specified.

\textbf{Harness Abstraction and Editable Interfaces.}
The editable harness space is organized according to the six functional coordinates introduced in Section~\ref{3.3HarnessEvolution}. To instantiate this abstraction in the agent framework, each coordinate is aligned with the concrete framework interfaces that control the corresponding functionality, such as skill and tool configuration, instruction prompts, action logic, observation processing, memory handling, and termination behavior. This alignment determines where each proposed harness modification is applied in the underlying agent system, as summarized in Table~\ref{tab:harness_coordinates}. All methods start from the same initial harness $H_0$ for each benchmark and backbone model. The benchmark environment, task specification, tool APIs, and evaluation protocol remain fixed throughout evolution.

\textbf{Prompt Templates.} We provide the key prompt templates used in the co-evolution procedure. The trajectory-analysis prompt is shown in Figure~\ref{fig:rubric_failure_trajectory}, the rubric proposal and outcome-blind judging prompts in Figures~\ref{fig:rubric_proposal_prompt}–\ref{fig:rubric_outcome-blind_judge}, and the harness intervention planning prompt in Figure~\ref{fig:harness_intervention_planning}. 

\begin{table}[t]
\centering
\small
\renewcommand{\arraystretch}{1.08}

\begin{tabular*}{\columnwidth}{
@{\extracolsep{\fill}}
l c c c
@{}
}
\toprule

\textbf{Domain}
& \makecell[c]{\textbf{Evol.}\\[-1pt]\scriptsize(Held-in)}
& \textbf{Val.}
& \makecell[c]{\textbf{Test}\\[-1pt]\scriptsize(Held-out)}
\\

\midrule

Information Retrieval
& 100 & 54 & 65 \\

Knowledge Work
& 60 & 36 & 50 \\

Software Engineering
& 45 & 26 & 36 \\

Code Implementation
& 128 & 54 & 86 \\

Math Reasoning
& 380 & 98 & 100 \\

\bottomrule
\end{tabular*}

\caption{Dataset statistics for the evolution, validation, and test splits across five evaluation domains.}
\label{tab:dataset_statistics}
\end{table}
\begin{table}[t]
\centering
\small
\setlength{\tabcolsep}{3.5pt}
\renewcommand{\arraystretch}{1.08}

\begin{tabularx}{\columnwidth}{
@{}
>{\raggedright\arraybackslash}p{0.18\columnwidth}
>{\raggedright\arraybackslash}p{0.29\columnwidth}
>{\raggedright\arraybackslash}X
@{}
}
\toprule

\multicolumn{1}{c}{\textbf{Coordinate}}
&
\multicolumn{1}{c}{\textbf{Functional Scope}}
&
\multicolumn{1}{c}{\textbf{Editable Content}}
\\

\midrule

Equip
& Capability provisioning
& Skills, tools, and reusable modules \\

Instruct
& Behavioral guidance
& System and task instructions \\

Act
& Action execution
& Tool-use and action strategies \\

Observe
& Observation processing
& Parsing, filtering, and summarization \\

Remember
& Context management
& Memory and context handling \\

Terminate
& Completion control
& Stopping and finalization \\

\bottomrule
\end{tabularx}

\caption{Unified harness coordinates, their functional scopes, and editable contents.}
\label{tab:harness_coordinates}
\end{table}

\subsection{Details of Diagnostic Utility}
\label{app:diagnostic_utility}

In this subsection, we provide the detailed definitions of the five criterion-level statistics used to compute the diagnostic utility in Section~\ref{3.2RubricEvolution}.

For each criterion $c$, the outcome-blind judge assigns every trajectory one of three labels: \textsc{Violated}, \textsc{Satisfied}, or \textsc{N/A}. Let $v_{\mathrm{pass}}$, $s_{\mathrm{pass}}$, and $n_{\mathrm{pass}}$ denote the numbers of violated, satisfied, and non-applicable trajectories among successful tasks. Similarly, let $v_{\mathrm{fail}}$, $s_{\mathrm{fail}}$, and $n_{\mathrm{fail}}$ denote the corresponding counts among unsuccessful tasks. Based on these statistics, the five components are defined as follows.

\textit{Discrimination ($D$):}
Measures how well a criterion distinguishes unsuccessful trajectories from successful ones by comparing its violation rates across the two groups:
\begin{equation}
D(c)
=
\frac{v_{\mathrm{fail}}}
     {v_{\mathrm{fail}}+s_{\mathrm{fail}}}
-
\frac{v_{\mathrm{pass}}}
     {v_{\mathrm{pass}}+s_{\mathrm{pass}}}.
\end{equation}

\textit{Coverage ($C$):}
Measures the proportion of unsuccessful trajectories for which the criterion identifies a violation:
\begin{equation}
C(c)
=
\frac{v_{\mathrm{fail}}}
     {v_{\mathrm{fail}}+s_{\mathrm{fail}}+n_{\mathrm{fail}}}.
\end{equation}

\textit{Predictive Validity ($P$):}
Measures whether observing a criterion violation increases the likelihood of task failure relative to the criterion's applicable population:
\begin{equation}
P(c)
=
\frac{v_{\mathrm{fail}}}
     {v_{\mathrm{fail}}+v_{\mathrm{pass}}}
-
\frac{v_{\mathrm{fail}}+s_{\mathrm{fail}}}
     {v_{\mathrm{pass}}+s_{\mathrm{pass}}
      +v_{\mathrm{fail}}+s_{\mathrm{fail}}}.
\end{equation}
Equivalently,
\begin{equation}
P(c)
=
\Pr(\mathrm{fail}\mid\mathrm{violated})
-
\Pr(\mathrm{fail}\mid\mathrm{applicable}).
\end{equation}

\textit{Novelty ($N$):}
Measures the proportion of failed trajectories uniquely identified by criterion $c$. Let $\mathcal{U}_c$ denote the set of failed trajectories violated only by criterion $c$, and let $N_F$ denote the total number of failed trajectories:
\begin{equation}
N(c)
=
\frac{|\mathcal{U}_c|}{N_F}.
\end{equation}

\textit{Redundancy ($\mathrm{Red}$):}
Measures the behavioral overlap between criterion $c$ and the remaining criteria. Let $\mathcal{V}(c)$ denote the set of trajectories violated by criterion $c$:
\begin{equation}
\mathrm{Red}(c)
=
\max_{c' \neq c}
\operatorname{Jaccard}
\left(
\mathcal{V}(c),
\mathcal{V}(c')
\right).
\end{equation}
\subsection{Detail of Rubric Admission and Lifecycle}\label{app:rubric_lifecycle}

In this subsection, we provide the detailed admission criteria and lifecycle transition rules used to maintain the evolving rubric pool introduced in Section~\ref{3.2RubricEvolution}.

\textbf{Birth Check.}
Before entering the maintained rubric pool, each newly discovered criterion undergoes a birth check designed to remove criteria that are unlikely to provide useful diagnostic signals. We reject a criterion under any of the following conditions:
(i) \textit{never applies}, when more than 80\% of its assessments are \textsc{N/A};
(ii) \textit{never fires}, when it produces no violations on unsuccessful trajectories; or
(iii) \textit{fires on all}, when its violation rate exceeds 90\% overall and also exceeds 90\% on successful trajectories.
These checks remove criteria that are respectively inapplicable, insensitive to actual failures, or insufficiently discriminative. Criteria that pass the birth check enter the lifecycle in the \textit{probation} state. Since lifecycle transitions are evaluated within the same evolution round, a newly admitted criterion may already appear as \textit{active} in the round-end visualization if it immediately satisfies the promotion condition.

\textbf{Lifecycle Transition Mechanism.}
Each admitted criterion is maintained in one of four states: \textit{probation}, \textit{active}, \textit{dormant}, or \textit{retired}. State transitions are evaluated after each evolution round according to the criterion's observed diagnostic statistics.

\textit{Probation.}
A newly admitted criterion first enters the \textit{probation} state. It is promoted to \textit{active} once it exhibits positive predictive validity ($P>0$). If it remains untriggered for four consecutive rounds, it is instead moved directly to \textit{retired}.

\textit{Active.}
An \textit{active} criterion is eligible for frontier construction. It becomes \textit{dormant} when its trigger rate remains below 5\% for three consecutive rounds or when the predefined N/A-drift condition persists for three consecutive rounds. It is directly retired when its predictive validity remains below $0.10$ for three consecutive rounds.

\textit{Dormant.}
A \textit{dormant} criterion is retained for monitoring but excluded from frontier construction. It is reactivated when its trigger rate rises to at least 15\%. If it remains dormant for four consecutive rounds without reactivation, it is moved to \textit{retired}.

\textit{Retired.}
The \textit{retired} state is terminal. Retired criteria no longer participate in criterion prioritization or frontier construction. Table~\ref{tab:rubric_lifecycle_rules} summarizes all lifecycle transition rules.

\begin{table}[t]
\centering
\small
\setlength{\tabcolsep}{3.5pt}
\renewcommand{\arraystretch}{1.08}

\begin{tabular}{@{}p{0.23\columnwidth}
                p{0.50\columnwidth}
                p{0.19\columnwidth}@{}}
\toprule
\textbf{Current State} & \textbf{Condition} & \textbf{Next State} \\
\midrule

Probation
& $P > 0$
& Active \\

Probation
& Untriggered for 4 rounds
& Retired \\

Active
& Trigger rate $< 5\%$ for 3 rounds
& Dormant \\

Active
& N/A drift for 3 rounds
& Dormant \\

Active
& $P < 0.10$ for 3 rounds
& Retired \\

Dormant
& Trigger rate $\geq 15\%$
& Active \\

Dormant
& Remains dormant for 4 rounds
& Retired \\

\bottomrule
\end{tabular}

\caption{Transition rules for the rubric lifecycle. Trigger rate is the violation frequency of a criterion, and N/A drift reflects a sustained rise in its not-applicable rate.}
\label{tab:rubric_lifecycle_rules}
\end{table}
\begin{figure*}[t]
    \centering
    \includegraphics[width=\linewidth]{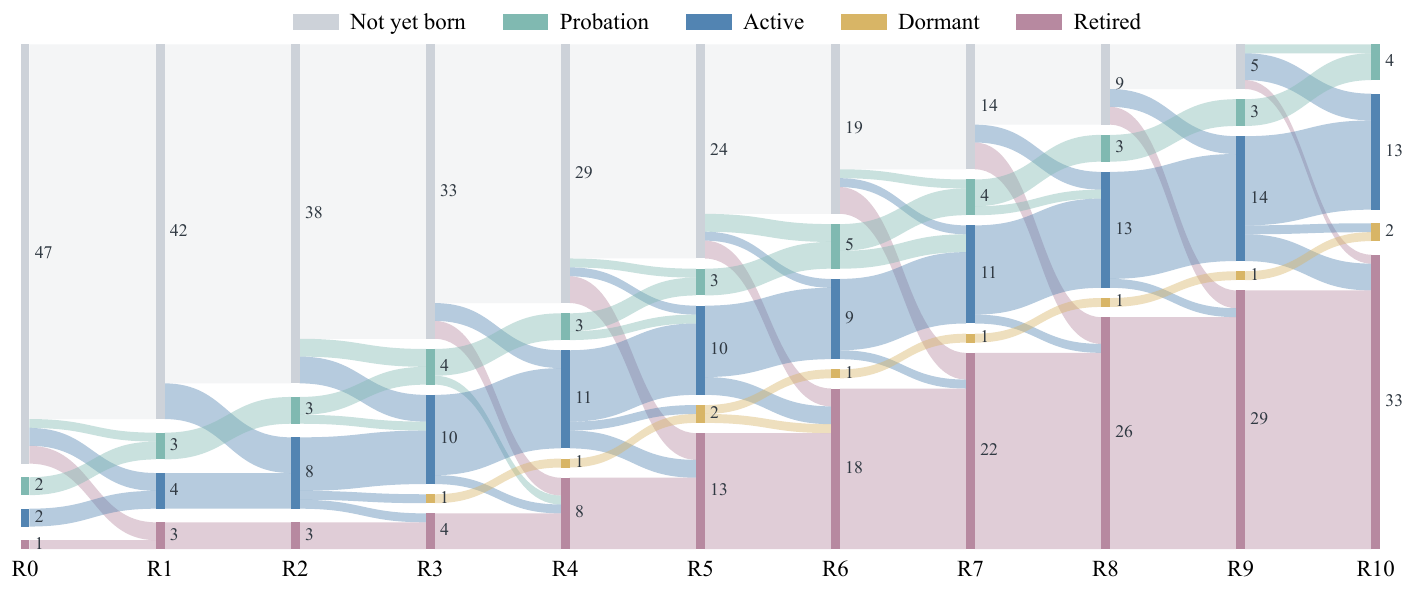}
    \caption{Rubric lifecycle across evolution rounds on the Qwen3.5-35B-A3B model with the knowledge work task.
    Each column shows criterion states at the end of a round,
    with flow widths proportional to the number of criteria.
    ``Not yet born'' denotes criteria introduced in later rounds;
    ``Retired'' is cumulative and includes candidates rejected at birth.}
    \label{fig:rubric_lifecycle}
\end{figure*}

\textbf{Visualization of Lifecycle Dynamics.}
Figure~\ref{fig:rubric_lifecycle} visualizes the lifecycle dynamics on Qwen3.5-35B-A3B for the Knowledge Work domain. Each column shows criterion states at the end of an evolution round, while the flows trace transitions between states. ``Not yet born'' denotes criteria introduced only in later rounds. For visualization, the \textit{retired} category also includes criteria rejected at birth as bookkeeping, although such criteria do not formally enter the lifecycle.

The figure reveals substantial turnover throughout evolution. New criteria continue to enter the lifecycle and become active as they satisfy the promotion condition, while previously active criteria move to dormant or retired states as their diagnostic relevance decreases. Dormant criteria may also return to the active set when the corresponding failures re-emerge. Consequently, the active criterion set is repeatedly reorganized rather than monotonically accumulated.

This turnover provides a mechanistic explanation for the frontier dynamics observed in the main analysis. As harness updates alter the behavioral distribution, some criteria lose diagnostic relevance while others become newly informative, continually reshaping the active set available for frontier construction.

\subsection{Details of Posterior Estimation and Conservative Ranking}
\label{app:posterior_ranking}

In this subsection, we provide the details of the conservative ranking strategy introduced in Section~\ref{3.2RubricEvolution}. Directly ranking criteria by their point-estimated utility can be unreliable when only a small number of trajectories support a criterion, since a few observations may produce an artificially high score. We therefore rank criteria using a lower posterior estimate of their diagnostic utility.

For each criterion, we construct a $2\times3$ contingency table over task outcome (\textit{pass} or \textit{fail}) and criterion assessment (\textsc{Violated}, \textsc{Satisfied}, or \textsc{N/A}). Since discrimination, coverage, and predictive validity are all computed from this same table, we estimate them jointly. We place a symmetric Jeffreys prior over the six cells, corresponding to adding $0.5$ to each observed count, and obtain a Dirichlet posterior over the joint outcome distribution.

We draw $B=2000$ samples from this posterior. For each sample, we recompute discrimination, coverage, and predictive validity, while keeping novelty and redundancy fixed because they are computed from trajectory-level set statistics. This produces a posterior distribution over the diagnostic utility. We then use its lower $q$-quantile as the conservative score:
\begin{equation}
\widetilde{U}(c)
=
\operatorname{Quantile}_{q}
\left(
\left\{
U^{(b)}(c)
\right\}_{b=1}^{B}
\right),
\end{equation}
where $q=0.2$ in all experiments.

This lower-quantile ranking favors criteria whose diagnostic value remains high under uncertainty, while reducing the influence of criteria supported by only a few observations. The resulting scores are used to rank active criteria, from which the Top-$K$ criteria are selected to form the rubric frontier.

\subsection{Robustness to Random Seeds}
\begin{table}[t]
\centering
\small
\renewcommand{\arraystretch}{1.10}

\begin{tabular*}{\columnwidth}{
@{\extracolsep{\fill}}
l
c
c
@{}
}
\toprule

\textbf{Method}
& \textbf{Held-in}
& \textbf{Held-out} \\

\midrule

\rowcolor{gray!12}
\multicolumn{3}{c}{
\textit{\textbf{Qwen3.5-35B-A3B on Knowledge Work}}
} \\

Vanilla Agent
& $51.1 \pm 1.0$
& $38.0 \pm 2.0$ \\

ACE
& $55.6 \pm 1.0$
& $42.0 \pm 2.0$ \\

EvoSkill
& $54.4 \pm 1.0$
& $41.3 \pm 1.2$ \\

Self-Harness
& $52.2 \pm 1.9$
& $41.3 \pm 3.1$ \\

AHE
& $55.0 \pm 1.7$
& $47.3 \pm 3.1$ \\

Meta-Harness
& $57.8 \pm 2.5$
& $38.0 \pm 2.0$ \\

\textbf{RuHarness}
& $\mathbf{58.9 \pm 1.0}$
& $\mathbf{50.0 \pm 2.0}$ \\

\midrule

\rowcolor{gray!12}
\multicolumn{3}{c}{
\textit{\textbf{GLM-4.7-Flash on Software Engineering}}
} \\

Vanilla Agent
& $31.1 \pm 2.2$
& $31.5 \pm 3.2$ \\

ACE
& $37.8 \pm 0.0$
& $45.3 \pm 1.6$ \\

EvoSkill
& $42.2 \pm 2.2$
& $39.8 \pm 1.6$ \\

Self-Harness
& $45.9 \pm 1.3$
& $45.4 \pm 3.2$ \\

AHE
& $42.9 \pm 1.3$
& $39.8 \pm 1.6$ \\

Meta-Harness
& $40.0 \pm 0.0$
& $38.0 \pm 1.6$ \\

\textbf{RuHarness}
& $\mathbf{48.9 \pm 2.2}$
& $\mathbf{51.9 \pm 1.6}$ \\

\bottomrule
\end{tabular*}

\caption{
Robustness to random seeds on two representative settings.
Results are reported as mean $\pm$ standard deviation over three independent runs.
}
\label{tab:seed_robustness}
\end{table}
To assess the robustness of different evolution methods to stochasticity, we repeat all methods with three independent random seeds on two representative settings: Qwen3.5-35B-A3B on Knowledge Work and GLM-4.7-Flash on Software Engineering. Dataset splits, initial harnesses, and all hyperparameters are kept fixed across runs. Table~\ref{tab:seed_robustness} reports the mean and standard deviation over the three runs.

RuHarness maintains the best average performance across both held-in and held-out evaluations under three independent random seeds. Importantly, the gains remain consistent across different random seeds, while the observed variance is comparable to that of competing methods. This trend is particularly clear on held-out tasks, where RuHarness maintains a clear advantage under both backbone--task combinations. These results indicate that the observed gains are robust to stochasticity in the evolution process.
\subsection{Effectiveness of Dynamic Evolution}
\label{app:dynamic_rubric_generalization}
\begin{figure}[t]
    \centering

    \begin{subfigure}[b]{0.49\columnwidth}
        \centering
        \includegraphics[width=\linewidth]{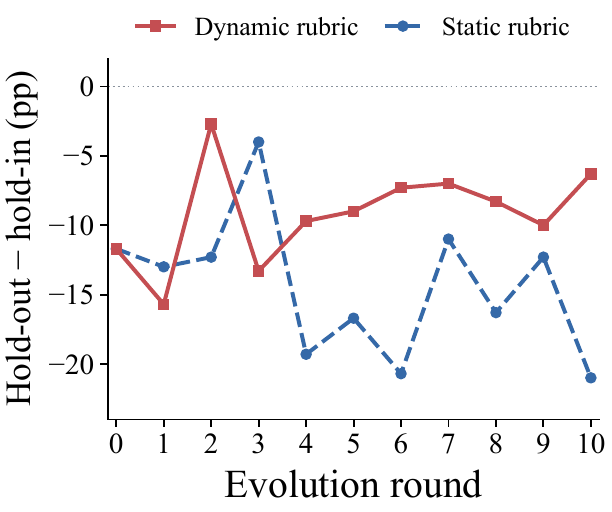}
        \caption{Generalization gap across evolution rounds.}
        \label{fig:dynamic_static_gap}
    \end{subfigure}
    \hfill
    \begin{subfigure}[b]{0.49\columnwidth}
        \centering
        \includegraphics[width=\linewidth]{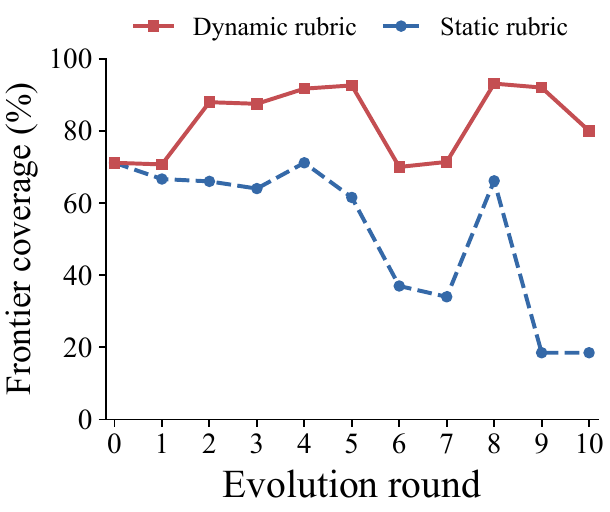}
        \caption{Frontier coverage across evolution rounds.}
        \label{fig:dynamic_static_coverage}
    \end{subfigure}

    \caption{
    Effectiveness of dynamic rubric evolution on the Qwen3.5-35B-A3B model with the knowledge work task.
    Compared with a static rubric, the dynamic rubric maintains a smaller held-out--held-in generalization gap and substantially higher frontier coverage over the course of evolution.
    }
    \label{fig:dynamic_rubric_effectiveness}
\end{figure}

In this subsection, we further analyze how dynamic rubric evolution affects generalization during harness optimization. We compare the full RuHarness framework with a static-rubric variant in which the rubric pool is fixed after initialization, while the remaining evolution procedure is kept unchanged. Figure~\ref{fig:dynamic_rubric_effectiveness} reports the round-wise generalization gap and frontier coverage on the Qwen3.5-35B-A3B Knowledge Work setting.

\textbf{Dynamic rubrics maintain a smaller generalization gap.}
We define the generalization gap as held-out performance minus held-in performance at each evolution round. As shown in Figure~\ref{fig:dynamic_static_gap}, although both variants can improve held-in performance during evolution, the static-rubric variant exhibits substantially larger fluctuations and increasingly negative gaps in later rounds. In contrast, the dynamic-rubric variant remains relatively stable across evolution and ends with a markedly smaller discrepancy between held-in and held-out performance. This suggests that harness optimization guided by an evolving rubric transfers more effectively to unseen tasks.

\textbf{Dynamic rubrics preserve coverage of current failures.}
Figure~\ref{fig:dynamic_static_coverage} further shows a clear divergence in frontier coverage, defined as the proportion of failed trajectories captured by at least one criterion in the currently prioritized rubric frontier. As evolution proceeds, the dynamic rubric maintains high coverage across most rounds, whereas the static rubric progressively loses coverage, particularly in later stages. This indicates that a fixed rubric becomes increasingly misaligned with the weaknesses exposed by the evolving harness, while continual rubric updates allow the frontier to track newly emerging failure patterns.

Taken together, these results provide additional evidence for the role of rubric evolution in continual harness optimization. A static rubric can still support improvements on the evolution tasks, but its diagnostic relevance gradually weakens as agent behavior changes. By continually updating and reprioritizing behavioral criteria, RuHarness preserves coverage of unresolved failures and consequently achieves more stable generalization throughout evolution.
\subsection{Effectiveness of Rubric Frontier}
\label{app:frontier_rubric}

\begin{figure}[t]
    \centering

    \begin{subfigure}[b]{0.49\columnwidth}
        \centering
        \includegraphics[width=\linewidth]{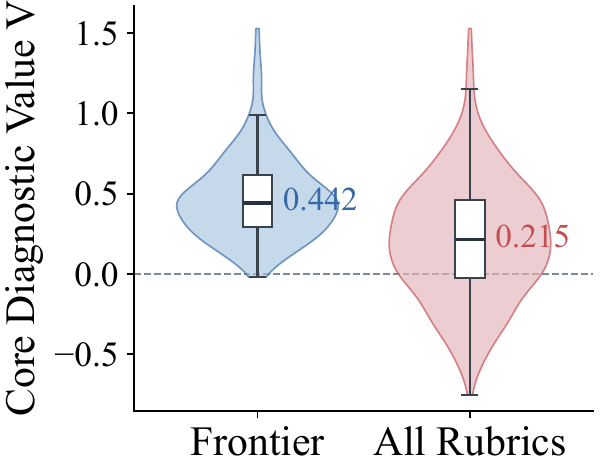}
        \caption{Distribution of core diagnostic value.}
        \label{fig:frontier_value_distribution}
    \end{subfigure}
    \hfill
    \begin{subfigure}[b]{0.49\columnwidth}
        \centering
        \includegraphics[width=\linewidth]{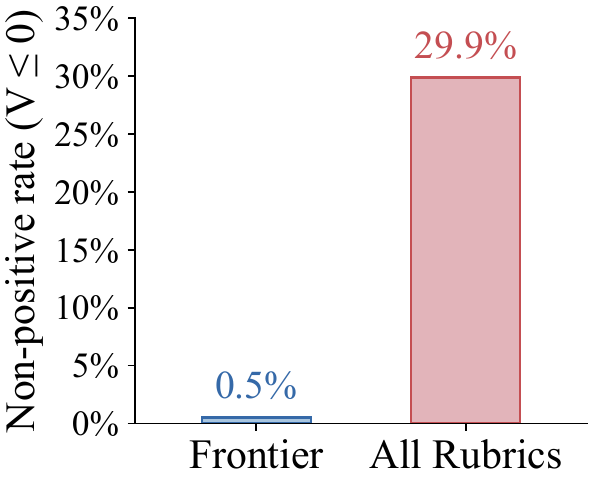}
        \caption{Rate of non-positive diagnostic value.}
        \label{fig:frontier_nonpositive_rate}
    \end{subfigure}

    \caption{
    Effectiveness of rubric frontier prioritization.
    Compared with the complete eligible rubric pool, the frontier concentrates criteria with substantially higher core diagnostic value and contains far fewer criteria with non-positive diagnostic value.
    }
    \label{fig:rubric_frontier_effectiveness}
\end{figure}

To understand why harness evolution benefits from a prioritized rubric frontier rather than the entire rubric pool, we compare the diagnostic composition of the frontier with that of all eligible rubrics. Figure~\ref{fig:rubric_frontier_effectiveness} reports the distribution of core diagnostic value and the proportion of criteria with non-positive diagnostic value. Here, the core diagnostic value is defined as the conservative posterior estimate:
\begin{equation}
V(c)=Q_{0.2}\!\left[D(c)+0.5C(c)+P(c)\right],
\end{equation}
which captures the principal diagnostic signals used for rubric prioritization.

\textbf{The frontier concentrates diagnostically valuable criteria.}
As shown in Figure~\ref{fig:frontier_value_distribution}, criteria in the frontier exhibit a substantially higher diagnostic-value distribution than the complete rubric pool. The median value increases from $0.215$ for all eligible rubrics to $0.442$ for the frontier. Since the full pool contains the frontier itself, this shift indicates that prioritization substantially increases the concentration of informative criteria presented to harness evolution, rather than simply reducing the number of criteria.

\textbf{Using all rubrics introduces substantial low-value guidance.}
Figure~\ref{fig:frontier_nonpositive_rate} further shows that only $0.5\%$ of frontier criteria have non-positive diagnostic value, compared with $29.9\%$ when considering the complete rubric pool. Thus, providing all rubrics to the proposer introduces a considerable fraction of weak diagnostic signals alongside the informative ones. This helps explain the degradation observed when frontier selection is removed: the frontier serves as a focused diagnostic filter that preserves high-value guidance while avoiding dilution by criteria that provide limited information about the agent's current weaknesses.
\subsection{Case Study of Rubric--Harness Co-Evolution}
\begin{table*}[t]
\centering
\small
\setlength{\tabcolsep}{4.0pt}
\renewcommand{\arraystretch}{1.15}

\begin{tabularx}{\textwidth}{
@{}
>{\raggedright\arraybackslash}p{0.05\textwidth}
>{\raggedright\arraybackslash}p{0.13\textwidth}
>{\raggedright\arraybackslash}X
>{\centering\arraybackslash}p{0.105\textwidth}
>{\raggedright\arraybackslash}X
@{}
}
\toprule

\textbf{Round}
& \makecell[l]{\textbf{Observed}\\\textbf{Pattern}}
& \textbf{Rubric Diagnosis}
& \makecell[c]{\textbf{Target}\\\textbf{Coordinate}}
& \textbf{Harness Response} \\

\midrule

\rowcolor{caseblue}
\multicolumn{5}{@{}l}{
\textbf{Case I: Knowledge Work --- From Generic Checking to Executable Verification}
} \\
\addlinespace[2pt]

R1
& \textit{Execution}
& Generated scripts should not be treated as valid merely because they were written;
\textcolor{rubricred}{\textbf{successful execution must be verified}}.
& \textsc{Observe}
& Require every generated script to be
\textcolor{harnessgreen}{\textbf{executed and explicitly checked for errors}}
before subsequent actions. \\

R3
& \textit{Error Analysis}
& Detecting a tool failure is insufficient if the
\textcolor{rubricred}{\textbf{underlying cause is not examined}}.
& \textsc{Observe}
& Require the agent to
\textcolor{harnessgreen}{\textbf{inspect the error output and identify its cause}}
before proceeding with unrelated commands. \\

R4
& \textit{Deliverable Check}
& Successful execution does not guarantee that the
\textcolor{rubricred}{\textbf{expected deliverable has actually been produced}}.
& \makecell[c]{\textsc{Terminate}\\+\textsc{Equip}}
& Add a completion-stage check that
\textcolor{harnessgreen}{\textbf{confirms the output file exists and matches the required filename}}.
\\

R5
& \textit{Programmatic Check}
& Declarative confirmation remains
\textcolor{rubricred}{\textbf{underspecified for reliable verification}}.
& \textsc{Terminate}
& Strengthen the check by requiring
\textcolor{harnessgreen}{\textbf{programmatic verification}}
of the output filename, e.g., through \texttt{os.path.basename}. \\

R8
& \textit{Pre-access Check}
& File validity should be established
\textcolor{rubricred}{\textbf{before the file is consumed downstream}}.
& \textsc{Remember}
& Move verification earlier by requiring
\textcolor{harnessgreen}{\textbf{file existence to be checked before reading}}
reference files. \\

\midrule

\rowcolor{caseyellow}
\multicolumn{5}{@{}l}{
\textbf{Case II: Information Retrieval --- From Candidate Retrieval to Auditable Verification}
} \\
\addlinespace[2pt]

R1
& \textit{Full-text Check}
& Search snippets provide only partial evidence and should not be sufficient to
\textcolor{rubricred}{\textbf{establish a candidate match}}.
& \makecell[c]{\textsc{Remember}\\+\textsc{Terminate}}
& Require the
\textcolor{harnessgreen}{\textbf{full document to be retrieved}}
before claiming that a candidate satisfies the target constraints. \\

R5
& \textit{Early Verification}
& The agent may form a
\textcolor{rubricred}{\textbf{premature match judgment}}
immediately after seeing a search result.
& \textsc{Observe}
& Explicitly mark search results as snippets and require
\textcolor{harnessgreen}{\textbf{full-text confirmation before accepting a match}}.
\\

R7
& \textit{Explicit Evidence}
& Verification remains difficult to audit when the response does not expose
\textcolor{rubricred}{\textbf{how individual constraints are satisfied}}.
& \makecell[c]{\textsc{Observe}\\+\textsc{Equip}\\+\textsc{Terminate}}
& Require the agent to
\textcolor{harnessgreen}{\textbf{state satisfied constraints and attach supporting evidence}}
for each claim. \\

R9
& \textit{Structured Audit}
& Individual evidence statements may still leave
\textcolor{rubricred}{\textbf{overall constraint coverage unclear}}.
& \makecell[c]{\textsc{Terminate}\\+\textsc{Remember}}
& Require a
\textcolor{harnessgreen}{\textbf{structured constraint-satisfaction matrix}}
and continue retrieval until all constraints are verified. \\

\bottomrule
\end{tabularx}

\caption{
Case studies of rubric--harness co-evolution with Qwen3.5-35B-A3B.
Across evolution rounds, diagnostic focus progressively shifts from coarse behavioral checks to more specific and operational requirements, which are translated into targeted modifications of the corresponding harness coordinates.
}
\label{tab:case_study}
\end{table*}
To examine how rubric evolution progressively reshapes harness optimization, we trace two representative evolution trajectories in Table~\ref{tab:case_study}. Rather than repeatedly targeting the same failure pattern, the diagnostic focus becomes increasingly specific across rounds, and the corresponding harness updates evolve from coarse checks to more operational, explicit, and structured interventions.

\textbf{Knowledge Work.}
The first trajectory exhibits a clear progression in verification requirements. It begins with a basic question of \emph{whether generated scripts are actually executed}, leading to an \textsc{Observe}-level rule that requires execution and error checking. The next stage moves beyond detecting failure to asking \emph{why the failure occurred}, prompting explicit analysis of returned error messages. The focus then shifts from execution correctness to \emph{deliverable correctness}: the agent must verify that the expected output artifact exists and matches the required filename. This requirement is subsequently strengthened from a verbal confirmation into \emph{programmatic verification}, making the check directly executable. In the final stage, verification is moved earlier in the workflow, from post-hoc confirmation to \emph{pre-access checking} before downstream file use. The trajectory therefore progresses from execution, to diagnosis, to artifact verification, to executable verification, and finally to proactive verification.

\textbf{Information Retrieval.}
The second trajectory shows a parallel progression from coarse evidence checking to structured auditing. It starts by requiring \emph{full-document retrieval} instead of relying on search snippets, establishing a stronger evidence basis for candidate matching. The next refinement moves this requirement closer to the retrieval step itself, preventing premature judgments immediately after observing a snippet. The focus then advances from merely verifying a candidate to making the verification \emph{explicit and auditable}, requiring the agent to state which constraints are satisfied and provide supporting evidence. Finally, the process becomes \emph{systematic}: individual evidence statements are organized into a constraint-satisfaction matrix so that all constraints are checked collectively rather than independently. Thus, the diagnostic trajectory evolves from source verification, to immediate verification, to explicit evidence grounding, and finally to structured constraint coverage.

These trajectories illustrate that rubric evolution progressively raises the granularity of behavioral supervision. As earlier requirements are incorporated into the harness, later rubrics shift attention to finer-grained weaknesses, which in turn induce more precise and structured modifications across the relevant harness coordinates. This creates a staged refinement process in which diagnosis and intervention become progressively more operational as evolution proceeds.

\clearpage
\onecolumn
\begin{tcblisting}{
    breakable,
    width=\columnwidth,
    title=\textbf{Failure Trajectory Analysis},
    colback=gray!5,
    colframe=black,
    colbacktitle=gray!10,
    coltitle=black,
    arc=1mm,
    boxrule=1pt,
    boxsep=0pt,
    left=2mm,
    right=2mm,
    top=2mm,
    bottom=2mm,
    before skip=4pt,
    after skip=4pt,
    toptitle=1.5mm,
    bottomtitle=1.5mm,
    listing only,
    listing options={
        basicstyle=\ttfamily\footnotesize,
        breaklines=true,
        breakautoindent=false,
        breakindent=0pt,
        columns=fullflexible,
        keepspaces=true,
        showstringspaces=false
    }
}
You are analyzing ONE agent trajectory to produce a factual failure report. Another system will read many of these reports together and design evaluation criteria from them. Your job is to record WHAT HAPPENED, precisely and verifiably -- not to judge the agent, and not to state principles.

VERIFIER RESULT (authoritative):
outcome: {outcome}
exception: {exception_type}
failed tests: {failed_tests}
verifier output (tail): {verifier_tail}
steps: {n_steps}
wall clock: {wall_clock_min} min
repeated-action ratio: {repetition_ratio}

TRAJECTORY:
{trajectory_text}

Produce ONE JSON object:
{
  "goal": "<what the task required, one sentence>",
  "final_state": "<where the run actually ended, observable facts only>",
  "divergence_step": <int>,
  "divergence_evidence": "<VERBATIM excerpt from that step: the actual action, command, and/or error text. Copy it; do not paraphrase.>",
  "failure_mode": "<exactly one of: {failure_modes}>",
  "agent_attributable": <true|false>,
  "attribution_note": "<one sentence>",
  "missed_opportunity": "<the specific alternative action available AT divergence_step -- name the action, tool, file, or operation the agent could have used instead. If none, write 'none'.>"
}

Rules:
- Every quote must appear verbatim above. Do not invent output.
- divergence_step is where progress STOPPED, usually much earlier than where the run ended.
- agent_attributable is false ONLY when no available action could have changed the result -- infrastructure broke, the agent harness malfunctioned, or the model API failed.
- An environment obstacle that the agent responded to badly is still attributable (true).
- missed_opportunity must be concrete and executable.
- "Should have been more careful", "should have verified the result", and "should have planned better" are rejected -- if that is all you can say, write "none".

Output ONLY the JSON object.
\end{tcblisting}
\vspace{-6pt}
\captionof{figure}{Prompt template for failure trajectory analysis.}
\label{fig:rubric_failure_trajectory}
\begin{tcblisting}{
    breakable,
    width=\columnwidth,
    title=\textbf{Rubric Proposal},
    colback=gray!5,
    colframe=black,
    colbacktitle=gray!10,
    coltitle=black,
    arc=1mm,
    boxrule=1pt,
    boxsep=0pt,
    left=2mm,
    right=2mm,
    top=2mm,
    bottom=2mm,
    before skip=4pt,
    after skip=4pt,
    toptitle=1.5mm,
    bottomtitle=1.5mm,
    listing only,
    listing options={
        basicstyle=\ttfamily\footnotesize,
        breaklines=true,
        breakautoindent=false,
        breakindent=0pt,
        columns=fullflexible,
        keepspaces=true,
        showstringspaces=false
    }
}
You are designing evaluation criteria for an agent rubric.

A criterion is one sentence describing an OBSERVABLE agent behavior. A judge will later read a trajectory and answer satisfied / violated / not_applicable against it. A criterion is only useful if that judge would reach DIFFERENT answers on failing runs and succeeding runs.

EXISTING RUBRIC POOL -- do not restate these:
{pool_summary}

FAILURE REPORTS ({n_fail} runs):
{failure_reports}

SUCCESS REPORTS ({n_pass} runs):
{success_reports}

BLIND SPOTS -- no existing criterion fired on these failures. Prioritize them:
{uncovered_list}

TOO BROAD -- these existing criteria fired on nearly every run this round, so they cannot separate good runs from bad ones. Where one names a real concern, write a sharper criterion that captures the same concern in a checkable way; it will replace the vague version. Where it names nothing real, ignore it:
{overfiring_list}

Propose {n} new criteria. Each must:

1. DISCRIMINATE -- describe behavior visible in the SUCCESS reports and rare or absent in the FAILURE reports. Derive it from the failures: read what went wrong, then write the action that would have gone right. Before you commit to a criterion, check it against the success reports: if a judge would mark it VIOLATED there too, it carries no information. Discard it and write a different one.

2. BE AFFIRMATIVE -- state what the agent DOES, never what it does not do. "The agent X", or "Before Y, the agent X". Never "the agent does not X", "the agent avoids X", or "the agent never X". A judge rules "satisfied" when the agent did the right thing, so the sentence has to name the right thing. If the failure you are describing is an action the agent should not take, write the action it should take instead.

A contrastive clause is fine and often sharper -- "targets a different mechanism RATHER THAN another variant of it" names a positive action and contrasts it. What is forbidden is a negated MAIN clause.

3. BE OBSERVABLE -- an action taken, a tool call issued, a check performed or skipped, or a response to an error. Not an internal state like "understands" or "is careful".

4. BE FALSIFIABLE -- a run must be able to clearly satisfy it. A criterion that nearly every run violates is worthless.

5. GENERALIZE WITHOUT GOING VAGUE -- no task-specific nouns. General does NOT mean abstract: prefer a concrete behavioral rule over a virtue.

6. NOT DUPLICATE the pool, or each other.

REJECTED -- unfalsifiable, fires on nearly every run:
"The agent validates its solution against task requirements before completion."
"The agent adapts its strategy when previous actions fail."

ACCEPTED -- same generality, but observable and discriminating:
"After the same action fails twice with the same error, the agent's next action targets a different mechanism rather than another variant of it."
"Before declaring completion, the agent executes or inspects the artifact it produced and verifies the resulting output."

REJECTED -- prohibitions. A judge can rule these "satisfied" for a run that did nothing at all, and they tell a reader modifying the harness what to remove rather than what to install:
"The agent does not issue more than three consecutive search actions without retrieving a document."
"The agent does not output a final answer while a constraint is unverified."

ACCEPTED -- the same two concerns, stated as the behavior that should happen:
"Within three search actions of starting a line of inquiry, the agent retrieves the full content of one of the results."
"Before outputting a final answer, the agent states, for each constraint in the goal, the retrieved evidence that satisfies it."

Output ONLY a JSON array of {n} strings. No markdown, no commentary:
[
  "First criterion sentence.",
  "Second criterion sentence.",
  ...
]
\end{tcblisting}
\vspace{6pt}
\captionof{figure}{Prompt template for rubric proposal.}
\label{fig:rubric_proposal_prompt}
\begin{tcblisting}{
    breakable,
    width=\columnwidth,
    title=\textbf{Outcome-Blind Rubric Judge},
    colback=gray!5,
    colframe=black,
    colbacktitle=gray!10,
    coltitle=black,
    arc=1mm,
    boxrule=1pt,
    boxsep=0pt,
    left=2mm,
    right=2mm,
    top=2mm,
    bottom=2mm,
    before skip=4pt,
    after skip=4pt,
    toptitle=1.5mm,
    bottomtitle=1.5mm,
    listing only,
    listing options={
        basicstyle=\ttfamily\footnotesize,
        breaklines=true,
        breakautoindent=false,
        breakindent=0pt,
        columns=fullflexible,
        keepspaces=true,
        showstringspaces=false
    }
}
You are evaluating whether the following criterion holds for an agent trajectory.

CRITERION:
{nl_description}

TRAJECTORY (task {task_id}):
{trajectory_text}

Judge the behavior described by the criterion on its own terms. Do not try to infer whether the run ultimately succeeded, and do not let a guess about that influence the verdict.

Verdicts:
"satisfied" - the trajectory shows the criterion being met
"violated" - the trajectory shows the criterion being broken
"not_applicable" - the situation the criterion describes never arose

Output ONLY a single JSON object with these fields:
{
  "verdict": "satisfied" | "violated" | "not_applicable",
  "reason": "<one-sentence justification>",
  "step_index": <0-based int of the offending step, or null if not violated>,
  "quoted_action": "<VERBATIM excerpt from that step, copied not paraphrased, or null if not violated>"
}

Do not include any other text - no markdown, no commentary. Just the JSON.
\end{tcblisting}
\vspace{6pt}
\captionof{figure}{Prompt template for outcome-blind rubric judging.}
\label{fig:rubric_outcome-blind_judge}
\begin{tcblisting}{
    breakable,
    width=\columnwidth,
    title=\textbf{Harness Intervention Planning},
    colback=gray!5,
    colframe=black,
    colbacktitle=gray!10,
    coltitle=black,
    arc=1mm,
    boxrule=1pt,
    boxsep=0pt,
    left=2mm,
    right=2mm,
    top=2mm,
    bottom=2mm,
    before skip=4pt,
    after skip=4pt,
    toptitle=1.5mm,
    bottomtitle=1.5mm,
    listing only,
    listing options={
        basicstyle=\ttfamily\footnotesize,
        breaklines=true,
        breakautoindent=false,
        breakindent=0pt,
        columns=fullflexible,
        keepspaces=true,
        showstringspaces=false
    }
}
You are designing interventions that make an agent satisfy rubric criteria it currently violates.

You plan at the level of the WHOLE harness, not one component at a time. A plan is one coordinated change: a theory of change, the criteria it targets, the trajectory evidence it was inferred from, and the components it edits - each with a distinct role. A plan that edits one component is a complete answer; a plan that edits three is only worth it when the evidence says the three have to move together.

THIS ROUND'S BEHAVIORAL FRONTIER - {n_criteria} rubric criteria that this round's runs still violate AND that best separate passing runs from failing ones. A judge rules each run "satisfied" or "violated" against each sentence; runs it rules "satisfied" pass substantially more often than runs it rules "violated". They are ordered by evidential value, NOT by how often they fire, so the first is not necessarily the most common or most severe problem:

{criteria_list}

{evidence_section}

MODIFIABLE COMPONENTS - these are the only things you can change:

{components_list}

Produce up to {n_candidates} plans. Each plan must:

- state a falsifiable hypothesis: what you believe is going wrong, and why these edits should stop it. "Improves quality" is not a hypothesis.
- cite the evidence it rests on, as concrete (task_id, step indices) pairs drawn from the runs above. Cite the steps that actually show the problem.
- name the criteria it targets, by id, from the frontier above.
- assign ONE ROLE PER COMPONENT, and never the same component twice in one plan. Roles must not duplicate or contradict each other.
- use at most {max_components} components. Fewer is usually right: every extra component is another way for the bundle to fail to load and another variable that cannot be attributed afterwards.
- state what observable behavior should change if the hypothesis is right, so a reader can tell later whether it was.

Plans must differ from each other in evidence, mechanism, target criteria, or component set. If you cannot construct {n_candidates} genuinely distinct plans, return fewer. A duplicate plan costs a full evaluation sweep and returns noise; an empty slot costs nothing.

Note on self-created obstacles: a common failure is a change that makes the agent's own job harder - for example, injecting control text into a query that only an external retrieval system sees, or adding a rule that blocks the very calls the agent needs. Before proposing an edit, ask what the agent loses by it.

Output ONLY a single JSON object:

{
  "plans": [
    {
      "hypothesis": "<falsifiable claim about the failing mechanism>",
      "target_criterion_ids": ["<criterion id from the frontier>", ...],
      "evidence": [
        {
          "task_id": "<task id exactly as shown above>",
          "step_indices": [<int>, ...],
          "explanation": "<what these steps show>"
        }
      ],
      "expected_behavior": "<what should observably change>",
      "component_roles": [
        {
          "component": "<component name>",
          "role": "<what this component does in the plan>"
        }
      ],
      "risks": "<how this plan could backfire>"
    }
  ]
}

Do not include any other text - no markdown, no commentary. Just the JSON.
\end{tcblisting}
\vspace{6pt}
\captionof{figure}{Prompt template for harness intervention planning.}
\label{fig:harness_intervention_planning}

\end{document}